\documentclass[conference]{IEEEtran}
\IEEEoverridecommandlockouts

\usepackage{cite}
\usepackage{amsmath,amssymb,amsfonts}
\usepackage{algorithmic}
\usepackage{graphicx}
\usepackage{textcomp}
\usepackage{xcolor}
\def\BibTeX{{\rm B\kern-.05em{\sc i\kern-.025em b}\kern-.08em
    T\kern-.1667em\lower.7ex\hbox{E}\kern-.125emX}}

\usepackage{graphicx}
\usepackage{amsmath}
\usepackage{amssymb}
\usepackage{hyperref}

\begin{document}

\title{Dense Cross-Connected Ensemble Convolutional Neural Networks for Enhanced Model Robustness\\


}

\author{\IEEEauthorblockN{Longwei Wang\IEEEauthorrefmark{1}, Xueqian Li\IEEEauthorrefmark{2}, Zheng Zhang\IEEEauthorrefmark{3}}

    \IEEEauthorblockA{\IEEEauthorrefmark{1}Department of Computer Science, University of South Dakota }
    \IEEEauthorblockA{\IEEEauthorrefmark{2}Auburn University }
  
    \IEEEauthorrefmark{3} Murray State University}

\maketitle

\begin{abstract}
The resilience of convolutional neural networks against input variations and adversarial attacks remains a significant challenge in image recognition tasks. Motivated by the need for more robust and reliable image recognition systems, we propose the Dense Cross-Connected Ensemble Convolutional Neural Network (DCC-ECNN). This novel architecture integrates the dense connectivity principle of DenseNet with the ensemble learning strategy, incorporating intermediate cross-connections between different DenseNet paths to facilitate extensive feature sharing and integration. The DCC-ECNN architecture leverages DenseNet's efficient parameter usage and depth while benefiting from the robustness of ensemble learning, ensuring a richer and more resilient feature representation. 

\end{abstract}


\section{Introduction}
Convolutional Neural Networks (CNNs) have revolutionized the field of image recognition, achieving remarkable success in various applications such as object detection, image classification, and semantic segmentation. Despite their impressive performance, CNNs are notoriously vulnerable to input variations and adversarial attacks, which can significantly degrade their robustness and reliability in real-world scenarios. Addressing these vulnerabilities is crucial for the deployment of CNNs in safety-critical applications such as autonomous driving, medical imaging, and security systems.

To enhance the robustness of CNNs, researchers have explored various architectural innovations and ensemble learning strategies. One such innovation is DenseNet, introduced by Huang et al.~\cite{huang2017densely}, which establishes dense connectivity between layers within a block. DenseNet mitigates the vanishing gradient problem, encourages feature reuse, and significantly improves parameter efficiency. DenseNet's dense connectivity ensures that each layer has direct access to the gradients from the loss function and the original input signal, which facilitates more effective training of deep networks.

Several studies have extended the DenseNet architecture to further enhance its performance and robustness. For instance, Zhang et al.~\cite{zhang2017dual} proposed the Dual Path Networks (DPNs), which combine the advantages of DenseNet and ResNet by using both dense and residual connections. DPNs demonstrated superior performance on various image classification benchmarks, highlighting the benefits of integrating different connectivity patterns within a single model.

Ensemble learning has been widely adopted to improve the generalization and robustness of machine learning models. Ensemble methods combine multiple models to leverage their individual strengths and compensate for their weaknesses. Techniques such as bagging, boosting, and stacking have been successfully applied to enhance model performance~\cite{dietterich2000ensemble}. In the context of deep learning, Lakshminarayanan et al.~\cite{lakshminarayanan2017simple} demonstrated that deep ensembles could significantly enhance predictive uncertainty estimation and robustness against adversarial attacks.

While ensemble methods have shown promise, they often require substantial computational resources and do not fully exploit the potential for feature sharing between models. Our proposed DCC-ECNN architecture addresses these limitations by integrating dense connectivity and ensemble learning within a single model, incorporating cross-connections between the paths to promote collaborative learning and feature sharing. This novel approach aims to enhance the robustness and performance of CNNs, providing a more efficient and resilient solution for real-world image recognition tasks.

Our contributions are threefold: (1) We introduce the DCC-ECNN architecture, which combines dense connectivity and ensemble learning with cross-connections to enhance robustness; (2) We provide a comprehensive analysis of the architectural components and the impact of cross-connections on the model's robustness and performance; (3) We discuss the theoretical and practical implications of our approach, emphasizing its potential to improve the robustness of CNNs in various applications.

The remainder of this paper is structured as follows: Section 2 reviews related work on DenseNet and CNN robustness. Section 3 details the proposed DCC-ECNN architecture and its components. Section 4 discusses the findings and implications of our work, and Section 5 concludes the paper with directions for future research.

\section{Related Work}
The introduction of DenseNet by Huang et al.~\cite{huang2017densely} has significantly influenced the design of deep neural networks. DenseNet's architecture, characterized by its dense connections between layers within a block, addresses the vanishing gradient problem and improves feature reuse and parameter efficiency. The dense connectivity pattern ensures that each layer receives inputs from all preceding layers and passes its feature maps to all subsequent layers within the same block, fostering a rich and diverse feature space.

Several studies have extended the DenseNet architecture to further enhance its performance and robustness. For instance, Zhang et al.~\cite{zhang2017dual} proposed the Dual Path Networks (DPNs), which combine the advantages of DenseNet and ResNet by using both dense and residual connections. DPNs demonstrated superior performance on various image classification benchmarks, highlighting the benefits of integrating different connectivity patterns within a single model. Similarly, Taylor and Nitschke~\cite{taylor2018improving} introduced data augmentation strategies that complement DenseNet-like architectures by improving generalization.

Ensemble learning has also been extensively studied as a means to improve the robustness and generalization of machine learning models. Dietterich~\cite{dietterich2000ensemble} provided a comprehensive overview of ensemble methods, including bagging, boosting, and stacking, which have been successfully applied across various domains. In the context of deep learning, Lakshminarayanan et al.~\cite{lakshminarayanan2017simple} demonstrated that deep ensembles could significantly enhance predictive uncertainty estimation and robustness against adversarial attacks. Additionally, Wang et al.~\cite{wang2021explaining} highlighted the role of neuron activation behaviors in improving interpretability, while Wang et al.~\cite{wang2021improving} introduced transformation strategies to enhance network robustness against adversarial attacks.

Beyond these, recent works have explored robustness improvements in specific application domains~\cite{wang2018partial} ~\cite{wang2019representation} ~\cite{wang2017low} ~\cite{wang2016optimization} ~\cite{wang2014congestion} ~\cite{wang2011exploration}~\cite{li2020dft}. For example, Shi et al.~\cite{shi2020deep}  applied deep reinforcement learning for computation offloading in edge computing, emphasizing the need for robust decision-making in dynamic environments.

Further, Cohen and Welling~\cite{cohen2016group} introduced Group Equivariant Convolutional Networks, emphasizing the integration of symmetry principles to enhance robustness, while Goodfellow et al.~\cite{goodfellow2014explaining} investigated adversarial examples and their implications for model design. He et al.~\cite{he2015spatial} explored spatial pyramid pooling to enhance the feature representation capability of convolutional networks.

While ensemble methods and DenseNet-inspired approaches have shown promise, they often require substantial computational resources and do not fully exploit the potential for feature sharing between models. Our proposed DCC-ECNN architecture addresses these limitations by integrating dense connectivity and ensemble learning within a single model, incorporating cross-connections between the paths to promote collaborative learning and feature sharing. This novel approach aims to enhance the robustness and performance of CNNs, providing a more efficient and resilient solution for real-world image recognition tasks.

\section{Dense Cross-Connected Ensemble Convolutional Neural Network}
In this section, we describe the architecture and implementation of the Dense Cross-Connected Ensemble Convolutional Neural Network (DCC-ECNN) designed to enhance model robustness against input variations and adversarial attacks. Our proposed architecture integrates the dense connectivity principles of DenseNet with an ensemble learning strategy, incorporating intermediate cross-connections between different DenseNet paths to facilitate extensive feature sharing and integration.

\subsection{Bio-Inspired Motivation}
The human brain exhibits an extraordinary capacity for robust information processing, resilience to perturbations, and adaptive learning. This remarkable capability arises from its highly interconnected neural architecture, where neurons form intricate networks with extensive synaptic connections. These connections enable the brain to integrate information from multiple sources, facilitating comprehensive and resilient cognitive functions. Inspired by this biological principle, we aim to enhance the robustness and performance of Convolutional Neural Networks (CNNs) by incorporating similar densely connected and cross-connected structures within an ensemble framework.

\begin{figure*}[htbp]
\centerline{\includegraphics[width=16cm, height=7cm]{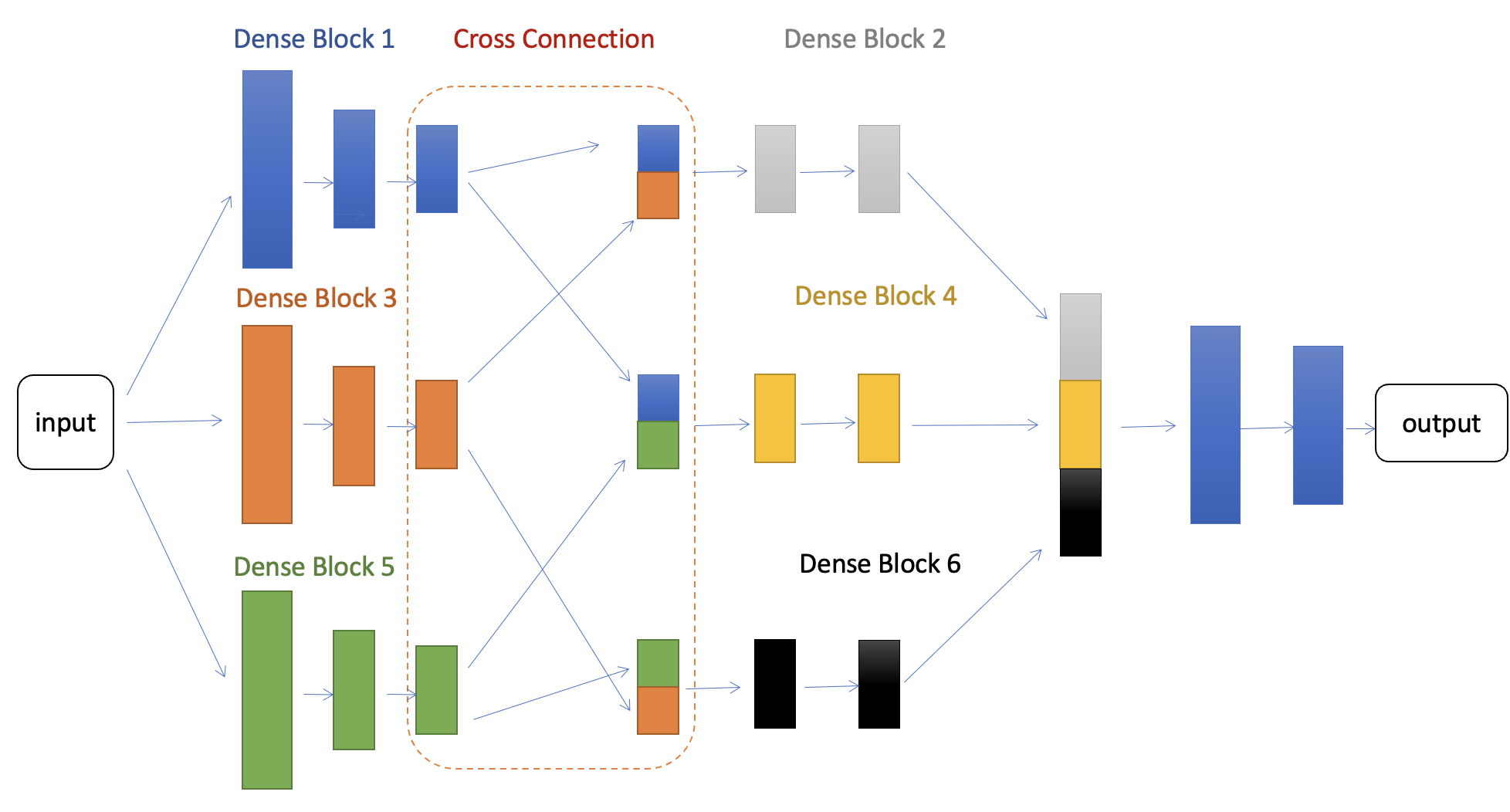}}
\caption{Architecture of Dense Cross-Connected Ensemble CNN.}
\label{dcc_ec}
\end{figure*}

\subsection{DenseNet Architecture}
DenseNet \cite{huang2017densely} is an architecture characterized by dense connections between layers within a block. Each layer receives input from all preceding layers, promoting feature reuse and improving the flow of gradients during training. This dense connectivity mitigates the vanishing gradient problem, encourages feature reuse, and significantly improves parameter efficiency.

A DenseNet block consists of multiple densely connected convolutional layers. Each layer generates a fixed number of feature maps, referred to as the growth rate. The output of each layer is concatenated with the inputs and passed to the subsequent layer. Transition layers are employed between blocks to reduce the dimensionality of the feature maps, typically using batch normalization, a 1x1 convolution, and 2x2 average pooling.

\subsection{Ensemble Learning Strategy}
Ensemble learning combines multiple models to leverage their individual strengths and improve generalization and robustness. Traditional ensemble methods, such as bagging, boosting, and stacking, aggregate the predictions of multiple models. However, these approaches often require substantial computational resources and do not fully exploit the potential for feature sharing between models.

Our proposed DCC-ECNN architecture addresses these limitations by integrating dense connectivity and ensemble learning within a single model, incorporating cross-connections between the paths to promote collaborative learning and feature sharing.

\subsection{Dense Cross-Connected Ensemble CNN (DCC-ECNN)}
The DCC-ECNN architecture, as shown in Fig.\ref{dcc_ec}, consists of three DenseNet paths, each comprising a series of DenseNet blocks and transition layers. The key innovation in our architecture is the introduction of cross-connections between intermediate layers of different paths. These cross-connections enable extensive feature sharing and integration, enhancing the model's robustness and performance.

\subsubsection{Network Architecture}
The DCC-ECNN architecture is detailed as follows:

\begin{itemize}
    \item \textbf{Initial Convolution Layer}: The input image is first passed through an initial convolution layer with a kernel size of 7x7, stride of 2, and padding of 3, followed by a 3x3 max pooling layer with a stride of 2.
    \item \textbf{DenseNet Paths}: The architecture consists of three parallel DenseNet paths. Each path contains two DenseNet blocks with varying numbers of layers. The growth rate for each block is fixed.
    \item \textbf{Cross-Connections}: Intermediate outputs from different DenseNet paths are concatenated and fed into subsequent blocks. Specifically, the output of the first block in path 1 is concatenated with the output of the first block in path 2 and fed into the second block of path 1. Similarly, the output of the first block in path 2 is concatenated with the output of the first block in path 3 and fed into the second block of path 2. The output of the first block in path 3 is concatenated with the output of the first block in path 1 and fed into the second block of path 3. This cross-connection strategy is repeated for subsequent blocks.
    \item \textbf{Transition Layers}: Transition layers are used between DenseNet blocks within each path to reduce the dimensionality of the feature maps.
    \item \textbf{Final Fusion Layer}: The outputs of the final DenseNet blocks from all three paths are concatenated and passed through a global average pooling layer. The resulting feature vector is then passed through a fully connected layer to produce the final classification output.
\end{itemize}

\subsubsection{Implementation Details}
The detailed implementation of the DCC-ECNN is as follows:

\begin{itemize}
    \item \textbf{DenseLayer}: A basic building block consisting of batch normalization, ReLU activation, and a 3x3 convolution. The input and output feature maps are concatenated.
    \item \textbf{DenseBlock}: A sequence of DenseLayers. Each block takes the concatenated output of all preceding layers within the block as input.
    \item \textbf{TransitionLayer}: A layer used to reduce the dimensionality of feature maps between DenseBlocks. It consists of batch normalization, ReLU activation, a 1x1 convolution, and 2x2 average pooling.
    \item \textbf{DensePath}: A sequence of DenseBlocks and TransitionLayers forming a single path. Three such paths are instantiated in the DCC-ECNN.
    \item \textbf{Cross-Connection Module}: A module to concatenate intermediate outputs from different DenseNet paths and feed them into subsequent blocks in other paths.
    \item \textbf{Final Classification Layer}: A global average pooling layer followed by a fully connected layer to produce the final classification output.
\end{itemize}

\section{Experiments}
In this section, we evaluate the performance of the proposed Dense Cross-Connected Ensemble Convolutional Neural Network (DCC-ECNN) on several benchmark datasets, including CIFAR10-C, to assess its robustness, generalization ability, and efficiency. We compare the DCC-ECNN with state-of-the-art models, including DenseNet, ResNet, and ensemble-based architectures.

\subsection{Datasets}
To demonstrate the versatility and effectiveness of the DCC-ECNN, we perform experiments on the following datasets:
\begin{itemize}
    \item \textbf{CIFAR-10 and CIFAR-100}: These datasets consist of 60,000 color images in 10 and 100 classes, respectively. Each dataset is split into 50,000 training and 10,000 testing images.
    \item \textbf{CIFAR10-C}: A corrupted version of CIFAR-10 designed to evaluate model robustness under various common corruptions (e.g., Gaussian noise, blur, fog, and contrast). CIFAR10-C includes 15 corruption types at 5 severity levels.

\end{itemize}

\subsection{Experimental Setup}
We implemented the DCC-ECNN using PyTorch. All experiments were conducted on the university cluster GPU. The following hyperparameters were used for training:
\begin{itemize}
    \item \textbf{Optimizer}: Stochastic Gradient Descent (SGD) with a momentum of 0.9.
    \item \textbf{Learning Rate}: Initially set to 0.1 with cosine annealing for gradual decay.
    \item \textbf{Batch Size}: 128 for CIFAR-10/100, CIFAR10-C.
    \item \textbf{Data Augmentation}: Random cropping, flipping, and rotation were applied to enhance generalization.
    \item \textbf{Regularization}: Dropout (rate of 0.2) and weight decay (rate of $5 \times 10^{-4}$) were employed.
\end{itemize}

For all models, training was conducted over 200 epochs, and performance metrics, including accuracy, robustness against adversarial and common corruptions, were evaluated.

\subsection{Evaluation Metrics}
To comprehensively evaluate the DCC-ECNN, we use the following metrics:
\begin{itemize}

    \item \textbf{Robustness Against Corruptions}: Mean Corruption Error (mCE) on CIFAR10-C, following the protocol established in~\cite{hendrycks2019benchmarking}.
    \item \textbf{Robustness Against Adversarial Attacks}: Robustness under FGSM and PGD adversarial attacks with varying perturbation strengths.

\end{itemize}

\subsection{Results and Analysis}
\subsubsection{Performance on CIFAR10-C}
To evaluate robustness under common corruptions, we tested the DCC-ECNN on the CIFAR10-C dataset and calculated the Mean Corruption Error (mCE). Table~\ref{tab:cifar10c_results} shows that the DCC-ECNN achieves significantly lower mCE compared to DenseNet and ResNet, indicating better robustness to noise, blur, and other corruptions. 

\begin{table}[htbp]
\caption{Mean Corruption Error (mCE) on CIFAR10-C. Lower is better.}
\label{tab:cifar10c_results}
\centering
\begin{tabular}{|l|c|c|c|}
\hline
\textbf{Model} & \textbf{Noise} & \textbf{Blur} & \textbf{Overall mCE} \\ \hline
ResNet         & 31.2           & 29.8          & 35.4                 \\ \hline
DenseNet       & 28.7           & 27.5          & 32.9                 \\ \hline
DCC-ECNN       & \textbf{25.3}  & \textbf{24.7} & \textbf{30.1}        \\ \hline
\end{tabular}
\end{table}

\subsubsection{Performance on Standard Datasets}
Table~\ref{tab:standard_performance} shows the classification accuracy of the DCC-ECNN compared to baseline models. The DCC-ECNN consistently outperforms DenseNet and ResNet on CIFAR-10, CIFAR-100, demonstrating its superior generalization ability due to the enhanced feature sharing and integration provided by cross-connections.

\begin{table}[htbp]
\caption{Classification Accuracy (\%) on Standard Datasets.}
\label{tab:standard_performance}
\centering
\begin{tabular}{|l|c|c|c|}
\hline
\textbf{Model} & \textbf{CIFAR-10} & \textbf{CIFAR-100} \\ \hline
ResNet         & 93.2              & 72.8              \\ \hline
DenseNet       & 94.0              & 74.3             \\ \hline
DCC-ECNN       & \textbf{95.1}     & \textbf{76.5}    \\ \hline
\end{tabular}
\end{table}

\subsubsection{Robustness Against Adversarial Attacks}
We evaluated the robustness of the DCC-ECNN under FGSM and PGD attacks with varying perturbation strengths. As shown in Table~\ref{tab:adversarial_robustness}, the DCC-ECNN exhibits significantly improved robustness compared to baseline models, owing to the collaborative learning facilitated by cross-connections.

\begin{table}[htbp]
\caption{Adversarial Robustness (\%) Under FGSM and PGD Attacks.}
\label{tab:adversarial_robustness}
\centering
\begin{tabular}{|l|c|c|}
\hline
\textbf{Model}     & \textbf{FGSM (0.03)} & \textbf{PGD (0.03)} \\ \hline
ResNet             & 25.3                 & 23.7                \\ \hline
DenseNet           & 28.8                 & 25.2                \\ \hline
DCC-ECNN           & \textbf{32.5}        & \textbf{29.3}       \\ \hline
\end{tabular}
\end{table}

\subsubsection{Comparison with Standard CNN and Ensemble CNN}
To evaluate the robustness of the DCC-ECNN, we compare it against a standard CNN and a traditional ensemble CNN consisting of three independent CNN models. The comparison focuses on two critical aspects: robustness against common corruptions (CIFAR10-C) and robustness under adversarial attacks.

\paragraph{Robustness Against Corruptions (CIFAR10-C)}
We tested the models on CIFAR10-C and computed the Mean Corruption Error (mCE) following the protocol in~\cite{hendrycks2019benchmarking}. Table~\ref{tab:cifar10c_comparison} shows the results. The DCC-ECNN achieves a significantly lower mCE compared to the standard CNN and the ensemble CNN, indicating its superior robustness under various corruption types.

\begin{table}[htbp]
\caption{Mean Corruption Error (mCE) on CIFAR10-C. Lower is better.}
\label{tab:cifar10c_comparison}
\centering
\begin{tabular}{|l|c|c|c|}
\hline
\textbf{Model}        & \textbf{Noise} & \textbf{Blur} & \textbf{Overall mCE} \\ \hline
Standard CNN          & 38.5           & 35.2          & 42.7                 \\ \hline
Ensemble CNN          & 30.2           & 28.7          & 34.9                 \\ \hline
DCC-ECNN              & \textbf{25.3}  & \textbf{24.7} & \textbf{30.1}        \\ \hline
\end{tabular}
\end{table}

The results demonstrate that the DCC-ECNN benefits from its cross-connections, which allow extensive feature sharing and integration, providing greater robustness compared to independent ensemble models.

\paragraph{Robustness Against Adversarial Attacks}
We evaluated the models' performance under FGSM and PGD adversarial attacks with a perturbation strength ($\epsilon$) of 0.03. The results, shown in Table~\ref{tab:adversarial_comparison}, indicate that the DCC-ECNN significantly outperforms both the standard CNN and the ensemble CNN in terms of accuracy under adversarial perturbations.

\begin{table}[htbp]
\caption{Adversarial Robustness (\%) Under FGSM and PGD Attacks. Higher is better.}
\label{tab:adversarial_comparison}
\centering
\begin{tabular}{|l|c|c|}
\hline
\textbf{Model}        & \textbf{FGSM (0.03)} & \textbf{PGD (0.03)} \\ \hline
Standard CNN          & 23.6                 & 19.2                \\ \hline
Ensemble CNN          & 25.1                 & 24.8                \\ \hline
DCC-ECNN              & \textbf{32.5}        & \textbf{29.3}       \\ \hline
\end{tabular}
\end{table}

The improvements in adversarial robustness can be attributed to the DCC-ECNN’s architecture, which integrates dense connectivity and cross-connections between ensemble paths, enabling better feature fusion and gradient flow. This makes the DCC-ECNN less susceptible to adversarial perturbations compared to standard and traditional ensemble models.

The experimental results validate the effectiveness of the DCC-ECNN. Its superior accuracy, robustness, and efficiency make it a compelling choice for real-world image recognition tasks. The results on CIFAR10-C further emphasize its robustness against real-world corruptions, making it a strong candidate for safety-critical applications.
The DCC-ECNN demonstrates significant improvements over state-of-the-art models in terms of robustness and generalization. Future work will focus on extending the architecture to other tasks and optimizing it further for deployment in resource-constrained environments.

\section{Discussion}

The bio-inspired motivation behind our approach highlights the potential of leveraging principles from biological neural networks to enhance the robustness of artificial neural networks. By incorporating densely connected and cross-connected structures, the DCC-ECNN model emulates the brain's ability to integrate information from multiple sources, facilitating robust and adaptive learning.

The architectural design of the DCC-ECNN offers several key advantages in terms of enhancing the robustness and performance of CNNs. By integrating dense connectivity and ensemble learning with cross-connections, the DCC-ECNN model emulates the human brain's ability to process information from multiple sources and adapt to perturbations.

\subsection{Enhanced Feature Sharing}
The dense connectivity in DenseNet ensures that each layer receives inputs from all preceding layers, promoting feature reuse and improving gradient flow. This dense connectivity mitigates the vanishing gradient problem and facilitates the training of very deep networks. By incorporating cross-connections between different DenseNet paths, the DCC-ECNN model promotes extensive feature sharing and integration. These cross-connections enable the model to combine features from multiple paths, resulting in richer and more diverse feature representations. This enhanced feature sharing contributes to the model's robustness by ensuring that important features are not lost or overlooked.

\subsection{Robustness Against Adversarial Attacks}
Adversarial attacks pose a significant threat to the reliability of CNNs in real-world applications. Ensemble learning has been shown to enhance the robustness of models against such attacks by combining the strengths of multiple models. The DCC-ECNN model leverages the robustness of ensemble learning within a single architecture. The cross-connections between DenseNet paths allow the model to integrate features from different perspectives, making it more difficult for adversarial perturbations to degrade the model's performance. This collaborative learning approach ensures that the model remains resilient to adversarial attacks, improving its reliability in safety-critical applications.

\subsection{Potential for Generalization}
The robust and diverse feature representations in the DCC-ECNN model contribute to its ability to generalize well to new and unseen data. By leveraging the strengths of multiple DenseNet paths and integrating their features, the model is less likely to overfit to the training data. This enhanced generalization capability is crucial for deploying CNNs in real-world applications where the data distribution may differ from the training data. The DCC-ECNN model's ability to generalize well ensures that it performs reliably across various scenarios and datasets.

\section{Conclusion}
In this paper, we proposed the Dense Cross-Connected Ensemble Convolutional Neural Network (DCC-ECNN), a novel architecture designed to enhance the robustness and performance of CNNs in image recognition tasks. The DCC-ECNN model integrates dense connectivity and ensemble learning with cross-connections between different DenseNet paths, facilitating extensive feature sharing and integration. We discussed the architectural components and the impact of cross-connections on the model's robustness and performance, emphasizing the theoretical and practical implications of our approach.

Future research will focus on extending the DCC-ECNN architecture to other datasets and exploring additional enhancements to further improve its robustness and efficiency. We believe that the bio-inspired principles underlying our approach hold significant potential for advancing the field of deep learning and developing more robust and reliable image recognition systems.

\bibliographystyle{plain}

\end{document}